# Robust Underwater Grasping of Sloped Objects with a Waterproof Passive Adaptive Gripper

Jooyoung Hong, Daewon Hong, and Joohyung Kim

***Abstract*— Robust grasping of everyday objects remains challenging for parallel-jaw grippers, particularly when handling sloped or asymmetric items that induce torque-driven rolling and shear slip. These challenges become even more severe in domestic environments such as kitchens, where objects are often wet or submerged, drastically reducing friction between the gripper and the object. To address these issues, we present a waterproof passive adaptive gripper that combines local and global adaptability for high-performance grasping in the submerged environments. The proposed gripper features a fully waterproof design that integrates a passive rotational joint for global self-alignment on sloped surfaces and passive variable-stiffness pads for local surface adaptation. Experiments in both dry and underwater conditions on various cylindrical and conical objects demonstrate superior holding capability and grasp robustness compared to rigid-pad and fixed-joint baselines. The proposed design offers a practical and robust solution that successfully enables stable underwater manipulation of diverse everyday objects, effectively addressing a critical gap in current underwater robotic grasping for daily-life applications.**

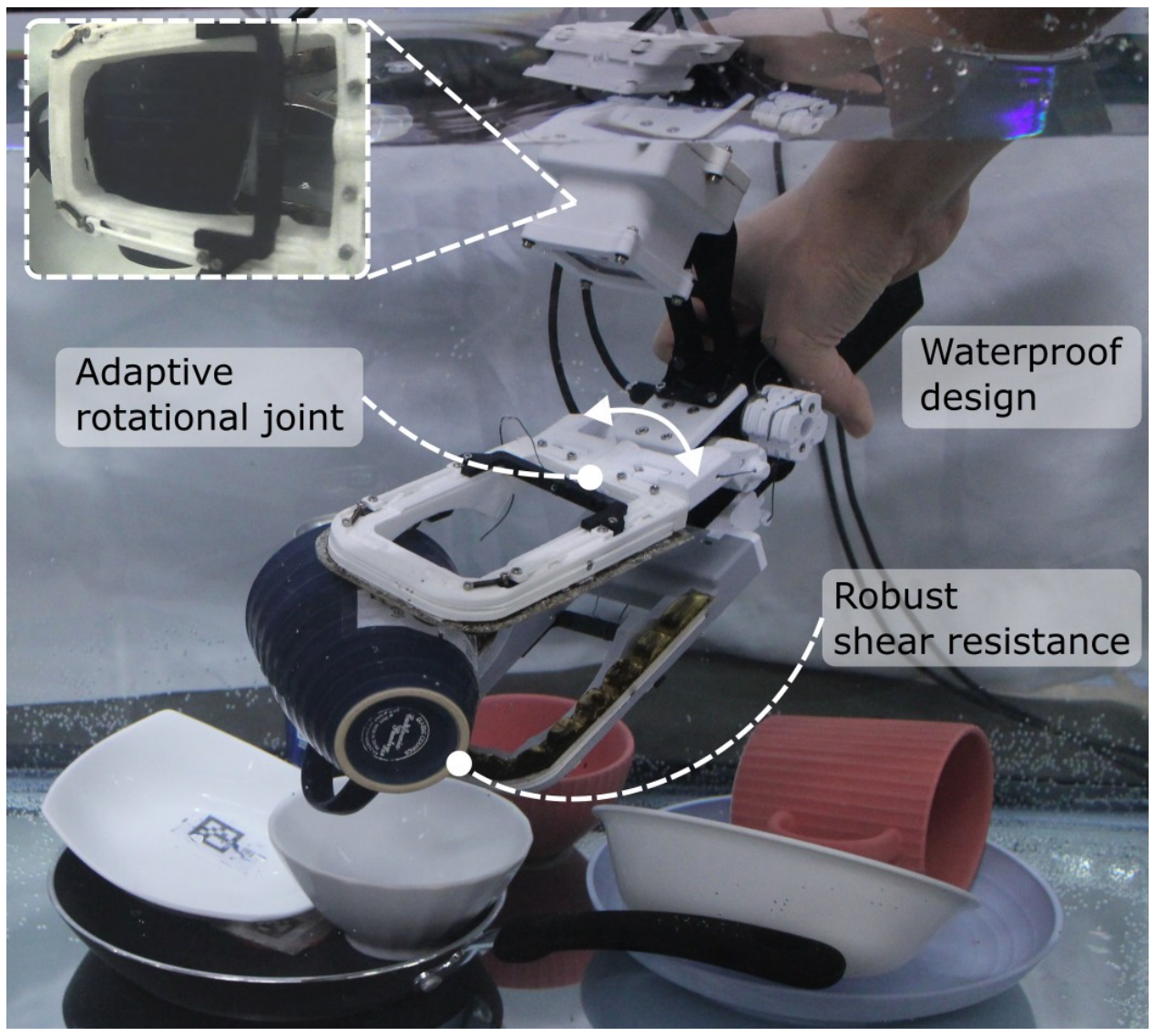

Fig. 1. The proposed waterproof passive adaptive gripper. The gripper passively self-aligns to conical objects through a rotational jaw and variable-stiffness pads, and achieves underwater grasp via waterproof integration.

## I. INTRODUCTION

Robust grasping of everyday objects is essential for service robots operating in domestic environments. In particular, automating chores such as dish collecting and dishwasher loading inherently requires interacting with items in wet or submerged conditions [1]–[4]. Handling these household items poses a profound challenge as their surfaces are often tapered, geometries asymmetric, and mass distributions off-center. In submerged environments, drastically reduced friction exacerbates these difficulties, such that sustained lifting under such low-friction conditions easily triggers rolling and tangential shear slip, leading to eventual grasp failure [5].

To improve grasp stability under such conditions, prior work has explored adaptive and soft grasping strategies, including Fin Ray-inspired fingers [6], underactuated hands [7], and various underwater soft or rigid-soft hybrid grippers [8]–[12] to improve stability. While these designs improve tolerance to geometric uncertainty through local adaptability, this local compliance alone cannot compensate for the macroscopic force misalignment that occurs on sloped, low-friction interfaces. Standard soft or adaptive fingertips merely deform locally against the surface while maintaining a fixed overall orientation, failing to neutralize the external offset torque.

Therefore, achieving stable grasp under such disturbances requires combining local compliance with global self-alignment, so that the gripper can both conform to local surface irregularities and reorient the overall contact geometry to reduce tangential shear and off-axis loading. Furthermore, stiffness increase is necessary to lock this aligned configuration and resist slip. Variable-stiffness grasping is highly effective for transitioning from a compliant to a load-bearing state [13], [14]. While jamming-based grippers have been adapted for underwater use [15]–[17], they mainly enhance local contact mechanics via compliance or variable stiffness. Yet local adaptation alone remains insufficient for robust grasping on sloped, low-friction surfaces. Moreover, many underwater jamming implementations rely on external vacuum, tubing, or additional sealing, which can increase system complexity.

To fill this gap, we present a waterproof passive adaptive gripper that combines local and global adaptability for robustly grasping sloped and asymmetric objects in both dry and underwater conditions (Fig. 1). The proposed gripper addresses the limitations by integrating: (1) a passive rotational joint that provides global adaptability by aligning the contact face with sloped geometries, thereby mitigating tangential shear and torque-driven rolling, (2) passive particle-jamming pads that provide local adaptability by conforming to surface irregularities before transitioning into a shear-resistant, load-bearing state under compression without external vacuum

All the authors are with the KIMLAB (Kinetic Intelligent Machine LAB), University of Illinois, Urbana-Champaign, IL 61801, USA {jh97, joohyung}@illinois.edu.

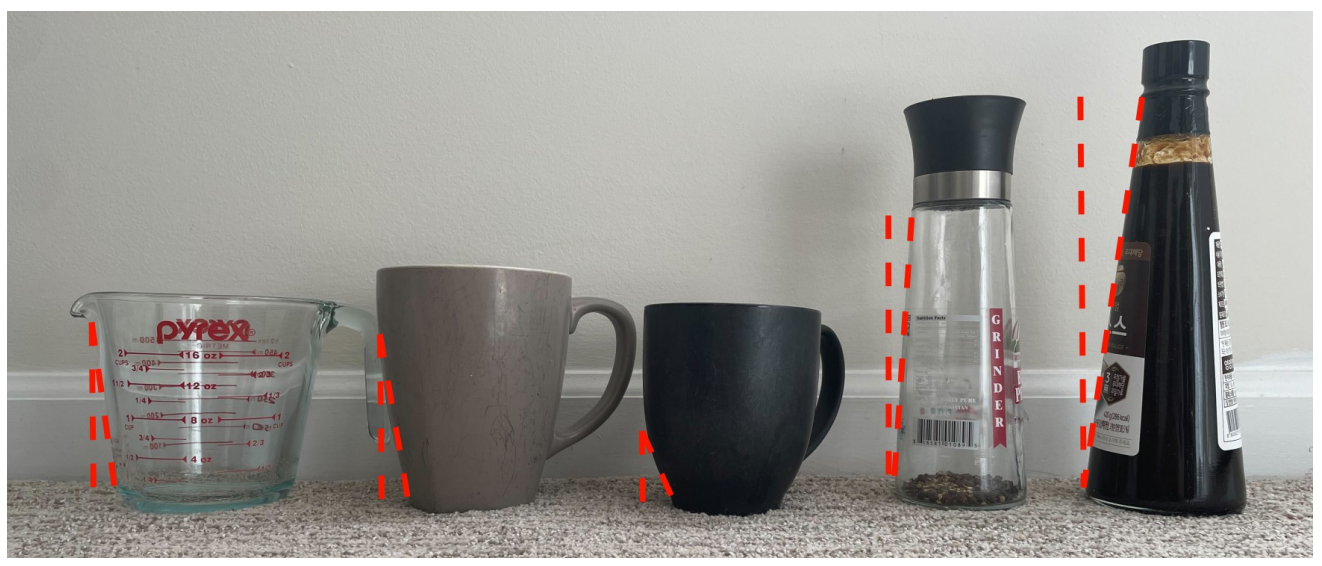

Fig. 2. Example of sloped household items.

TABLE I
SLOPE ANGLES OF HOUSEHOLD ITEMS

| Item | Half angle (deg) | Total angle (deg) |
|---|---|---|
| Measuring cup | 10 | 20 |
| Mug 1 | 8 | 16 |
| Mug 2 | 12 | 24 |
| Grinder | 8 | 16 |
| Sauce bottle | 11 | 22 |
| Plastic cup | 8 | 16 |
| Disposable cup | 7 | 14 |

systems, and (3) waterproof system integration that enables submerged operation of the passive adaptive mechanism. Experiments demonstrate that the proposed design improves holding capability and shear resistance compared to rigid-pad and fixed-joint baselines.

The rest of this paper is organized as follows. Section II introduces the design of the proposed gripper, detailing the mechanisms that enable its local and global adaptability. Section III presents a thorough analysis and experimental validation of the performance of the gripper in dry and submerged environments, alongside demonstrations of grasping various kitchenware under water.

## II. HARDWARE DESIGN

In this section, we present the design of the proposed gripper and its task-driven requirements for grasping sloped kitchen items. We then describe the key mechanisms that enable robust underwater grasping, including the passive rotational joint, the passive particle jamming pads, and the waterproof system integration.

### A. Sloped Kitchen Items

To inform the design requirements, we surveyed common kitchen and daily-life objects and measured representative surface slopes, as shown in Fig. 2. The object selection was task-driven, focusing on wet domestic manipulation scenarios where sloped or tapered kitchenware can induce rolling and shear slip during grasping. Many of these objects exhibit sloped or conical geometries, with maximum slope angles up to approximately 20°, as shown in Table I. Therefore, the passive rotational jaw was designed to accommodate target angles up to 30°, exceeding the measured slopes while providing tolerance for grasp pose error and geometric variation.

Beyond slope angle, the target objects also vary in diameter, surface material, and mass distribution, which together induce off-axis torques during lifting and pulling. In a standard parallel-jaw pinch grasp, the inability to conform to the sloped geometry often results in unstable point or line contacts. Consequently, such torques directly translate into tangential shear forces at the interface, causing progressive rolling, pose drift, or immediate slip even when the initial grasp appears stable.

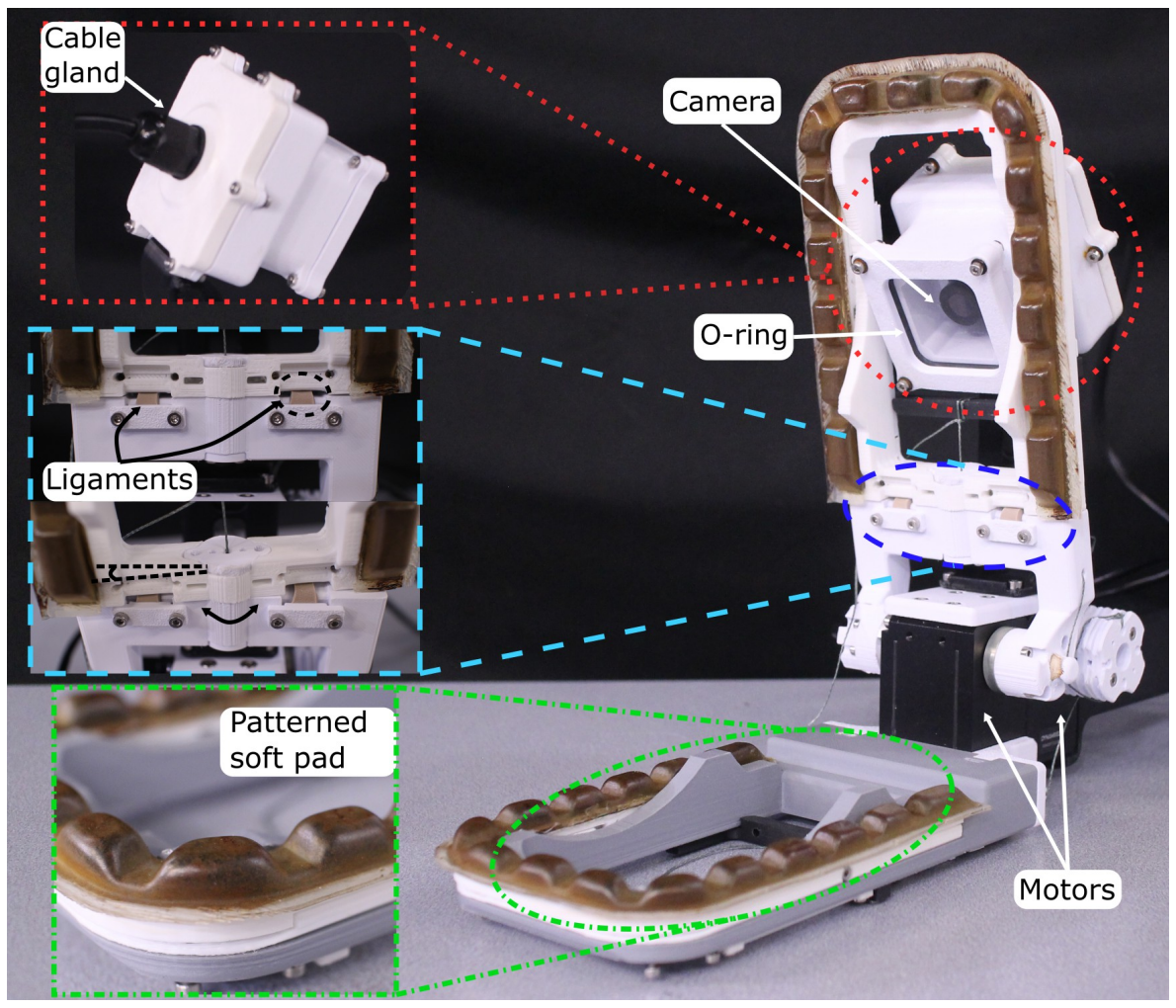

Fig. 3. Hardware overview of the proposed gripper. The design integrates a passive rotational joint for geometric self-alignment, passive particle jamming pads for enhanced shear resistance, and waterproof integration, enabling robust grasping in submerged environments.

### B. Design Overview

The proposed gripper, illustrated in Fig. 3, is designed to achieve robust grasping of sloped objects under submerged conditions. To accomplish this, the hardware is integrated based on three main design goals:

1) **Adaptive grasping by a passive rotational joint:** The jaw mechanism incorporates a passive rotational degree of freedom that allows the gripper to self-align with sloped surfaces. This accommodates varying surface curvatures and transitions unstable point contacts into stable surface contacts, effectively preventing shear slip.
2) **Enhanced shear resistance:** To resist shear slip and ensure stability after contact, the gripper utilizes passive particle jamming pads. During initial closure, the pad increases the contact area to reduce local pressure. On the other hand, under further compression, it provides a passive stiffness transition that securely holds the grasp by conforming to the shapes of objects.
3) **Waterproof integration:** For reliable underwater operation, the entire system is fully waterproof. It consists of a watertight camera housing and waterproofed actuation.

### C. Passive Rotational Joint

The passive rotational joint was introduced to enable self-alignment on sloped object surfaces. The joint is implemented using elastic ligaments that connect the jaw body to a rotating distal frame. To achieve the desired compliance and durability, we utilize medium-soft rubber strips (Shore 40A, 1.60 mm thickness, 20.7 MPa tensile strength, McMaster-Carr.) When the jaw first contacts a sloped surface, the contact normal does not generally align with the jaw closing direction. The passive joint allows the distal frame to rotate, thereby reducing tangential shear components at the interface. By reducing the angular mismatch between the jaw and the object surface, this passive rotation shifts the contact condition from localized edge or line contact toward distributed surface contact, which helps reduce shear-dominated loading and torque-driven rolling on sloped objects.

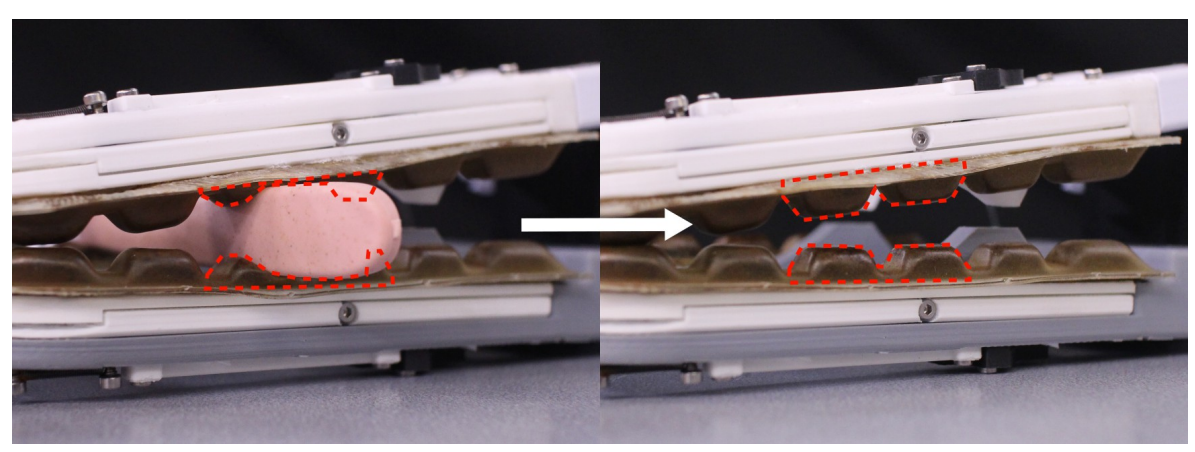

Fig. 4. Operating principle of the passive particle jamming pad. The pad conforms to the object during closure and relies on the elasticity of the flexible membrane to unjam and redistribute particles upon release.

The ligament stiffness provides two complementary functions: it generates a restoring torque that returns the jaw to a neutral pose when unloaded, and it provides compliant rotational accommodation during contact, enabling the jaw to settle into a mechanically stable configuration. In our design, the joint supports rotation up to approximately 30°, exceeding the measured slope angles of common household items to provide margin for irregular geometries and grasp perturbations. Once self-aligned, the established surface contact distributes external forces more effectively and helps prevent the off-axis rolling that typically causes standard parallel jaws to fail.

### D. Passive Particle Jamming Pad

The soft pads adopted the passive particle jamming mechanism to achieve a passive bi-directional stiffness transition from a compliant state to a load-bearing state under compression. The pad consists of a flexible Thermoplastic Urethane (TPU) outer membrane, filled with ground coffee particles. Two TPU membranes are heat-sealed together, so the pads are waterproof. During initial contact, the membrane and particles conform to the surface of the object, increasing the contact area and accommodating surface curvature. As compression increases, the particles tightly interlock, increasing inter-particle friction and forming a jammed structure highly resistant to deformation. This compression-induced stiffness transition enhances shear resistance and reduces slip after the grasp is established, following the passive jamming principle reported in previous passive particle jamming gripper studies [18], [19]. In this work, this mechanism is integrated with a waterproof membrane and a passive rotational joint to extend the same passive adaptation principle to submerged grasping of sloped objects.

Fig. 4 illustrates the shape adaptation and recovery cycle of the jamming pads. Upon contact, the pad compresses, and its outer membrane deforms to match the geometry of the object. Once the object is released, the pad passively recovers its initial shape, driven by the elastic restoring force of the TPU membrane. This passive shape recovery naturally unjams and redistributes the internal particles back to their original state, resetting the gripper for subsequent grasps without requiring external pneumatic systems.

To further enhance shear resistance in submerged conditions, the TPU membrane incorporates a macroscopic groove pattern. Under the submerged conditions, smooth contact interfaces often trap a fluid film, drastically reducing friction and causing shear slip. The grooves mitigate this effect by providing drainage channels that squeeze out trapped fluid during compression. Consistent with prior findings [20], [21], this surface texturing helps restore frictional margins and works synergistically with the internal jamming mechanism, ensuring robust torque resistance and stable grasping of slippery underwater objects.

### E. Waterproof Integration

The gripper is designed for submerged operation through waterproof actuation and a watertight camera enclosure. For actuation, off-the-shelf waterproof actuators (XW430-T333-R, ROBOTIS) were selected to avoid common failure modes of custom-sealed structures.

A compact USB camera (ELP-USBFHD01M-L21) is mounted on the gripper for visual observation. To enable underwater use, we designed a custom watertight housing that encloses the camera. A cable gland seals the cable pass-through while preserving flexibility and durability for repeated assembly and use.

The enclosure uses O-ring seals to provide repeatable, watertight assembly. O-rings are seated at both the front and rear interfaces. At the front, a transparent acrylic window compresses the O-ring against the housing, forming the seal while maintaining optical clarity for the camera.

## III. EXPERIMENTS

We evaluate the proposed gripper by measuring holding force in controlled pulling experiments conducted in both dry and submerged conditions to quantify the effect of underwater operation on holding capability. Each object was pulled 10 times, and the holding force was reported as the mean across trials.

### A. Experimental Setup

The pulling setup is shown in Fig. 5. The gripper was rigidly fixed on a frame placed inside a water tank as shown in Fig. 5(a), and the gripper holds objects with the same commanded current values to provide identical nominal actuator-level gripping conditions for all holding experiments. A motor mounted above the tank drove a pulley to pull the object via a low-stretch braided wire (PowerPro, Braided Spectra Fiber Microfilament line). As the motor reeled the wire, the object experienced a horizontal pulling force along the indicated direction in Fig. 5(a). The wire tension was converted to a compressive load on a compression load cell (FX-29, TE), and the peak force during each pull was recorded as the holding force. The pulling speed was set to a constant value of 7.0 mm/s to provide slow, repeatable, quasi-static loading and to minimize dynamic force transients associated with the pulling line and pulley system. For the underwater condition, the experiments were conducted in a static, clear-water tank to isolate the mechanical contribution of the gripper from external hydrodynamic disturbances.

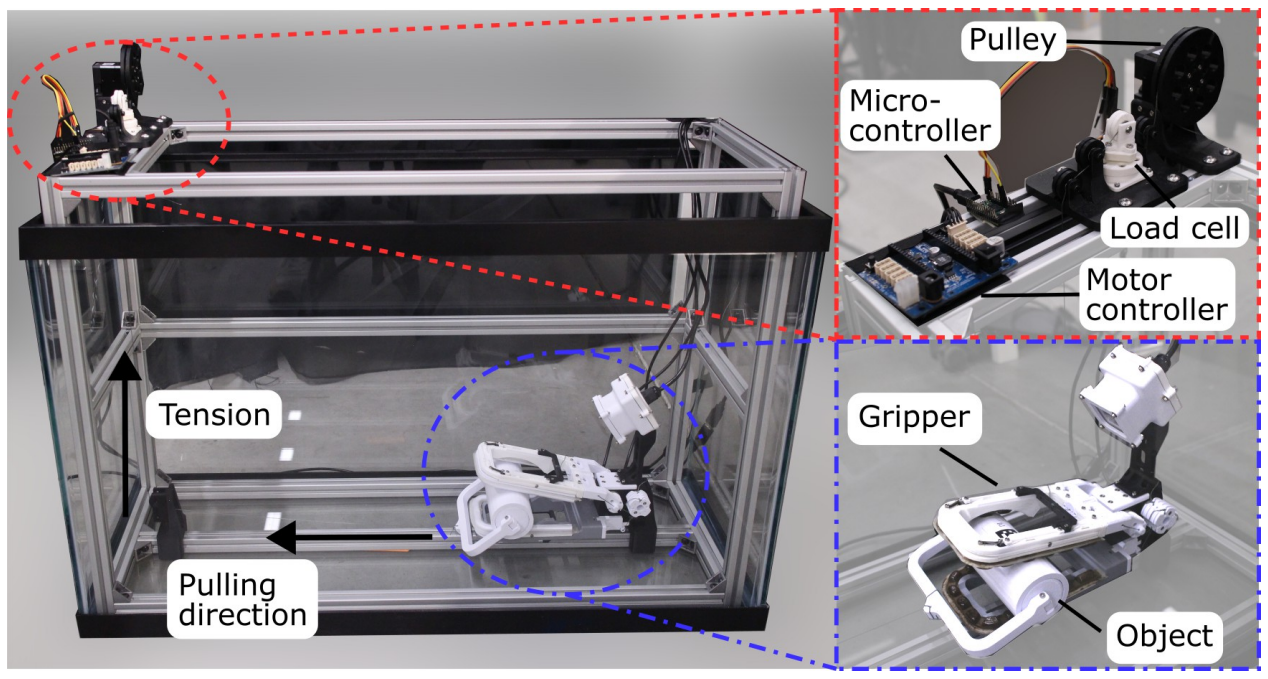

(a) Experimental setup for pulling tests

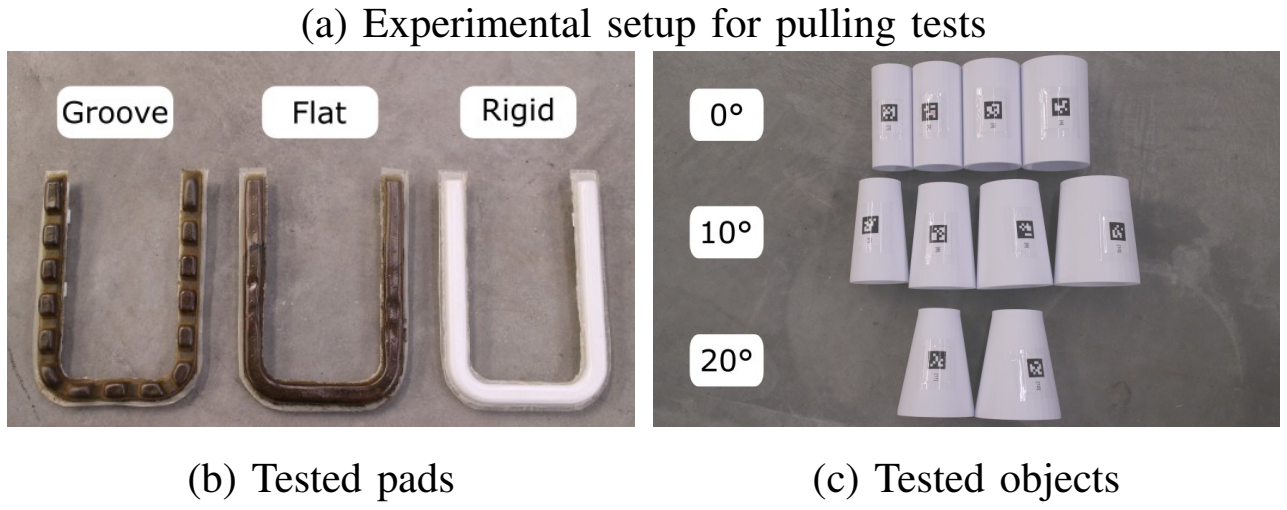

(b) Tested pads (c) Tested objects

Fig. 5. Experimental setup for evaluating holding force under dry and underwater conditions. (a) Overview of the pulling test conducted in a water tank. (b) The three tested pad types: grooved, flat, and rigid. (c) Cylindrical and conical test objects with varying diameters and slope angles.

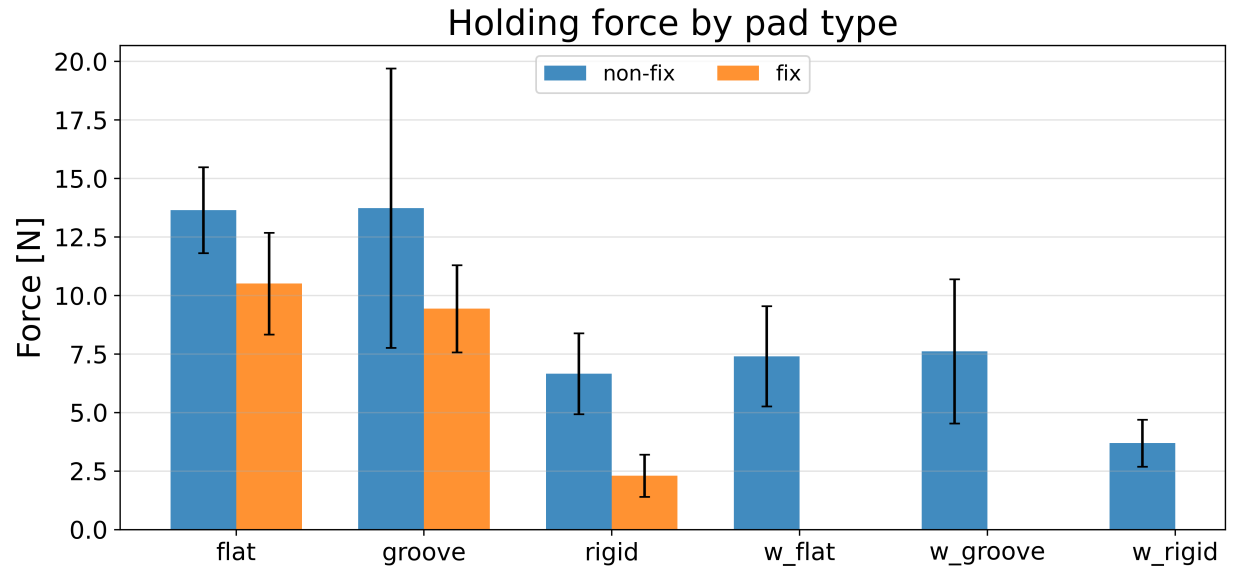

(a) Ablation study of design components

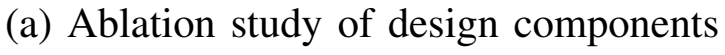

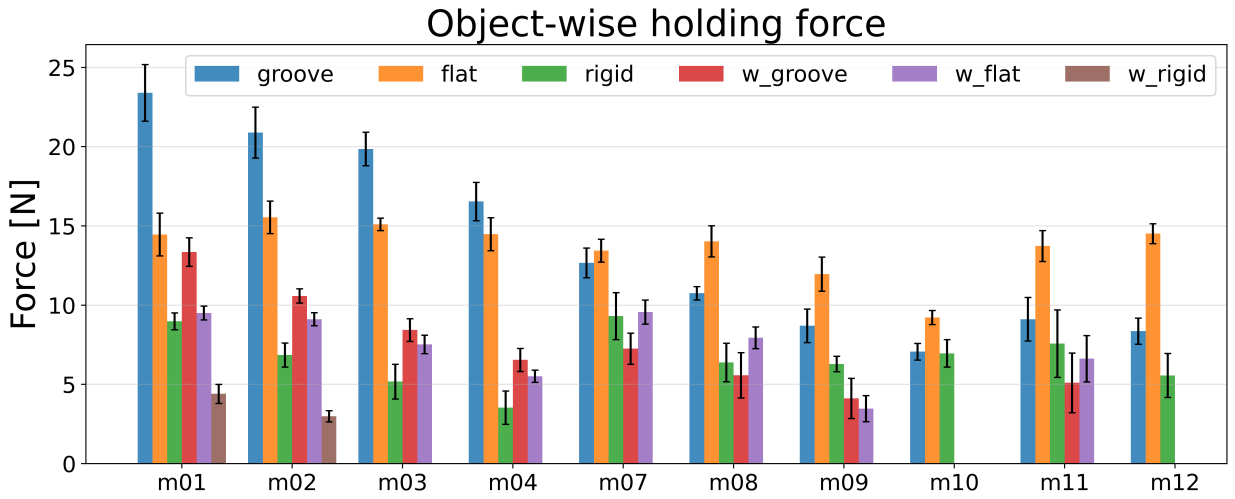

(b) Object-wise holding force comparison

Fig. 6. Holding force analysis across design variants and object geometries. Prefix 'w' indicates tested under water. (a) Comparison of the three contributions: the passive rotational joint (free vs. fixed), the particle jamming mechanism (jamming vs. rigid), and the surface pattern (grooved vs. flat). (b) Detailed performance breakdown for each target object, illustrating how diameter and slope angle influence the holding capacity in both dry and underwater environments across design variants.

To validate the contributions of surface patterning and passive particle jamming, we compared three pad types: grooved, flat, and rigid, as shown in Fig. 5(b). Both the grooved and flat pads were filled with ground coffee, whereas the rigid pad contained a 3D-printed PLA insert to serve as a non-jamming baseline. All pads were sealed using the same TPU outer membrane to ensure that the surface friction remained consistent across all conditions.

Fig. 5(c) shows the tested cylindrical and conical objects. Object IDs m1-m4 denote cylindrical objects with diameters of 50, 60, 70, and 90 mm, respectively; m7-m10 denote $10^\circ$ conical objects with nominal diameters of 50, 60, 70, and 90 mm, respectively; and m11-m12 denote $20^\circ$ conical objects with nominal diameters of 50 and 60 mm, respectively. Only two $20^\circ$ conical objects were used because the larger-side diameters exceeded the effective grasp range of the gripper.

### *B. Holding Force*

In this section, we evaluate holding capability across design variants to isolate the contributions of (i) the passive rotational joint, (ii) the particle jamming pad, and (iii) the pad surface pattern. We conducted 10 independent pulling trials per object, and the measured peak holding forces are summarized in Fig. 6. Note that in the submerged environment, the rigid pad succeeded only for the Marker 1 and 2. Moreover, all pad configurations failed for the largest conical objects at the steepest cone angles.

*1) Passive rotational joint:* To isolate the effect of the passive rotational joint, we compared the free-joint configuration against a fixed-joint baseline using a mechanical fixture. As shown in Fig. 6(a), the free-joint configuration achieved a higher overall mean holding force of 9.75 N compared to the fixed-joint baseline of 7.71 N, corresponding to a 26.5% improvement. This supports the hypothesis that passive self-alignment reduces tangential shear and prevents off-axis rolling by allowing the jaw to do self-alignment on sloped surfaces.

*2) Particle jamming pad:* Across all tested objects in dry conditions, the flat particle-jamming pad achieved substantially higher holding forces than the rigid baseline. The flat jamming pad produced 13.65 N, whereas the rigid pad produced only 6.66 N. These results demonstrate that the passive particle jamming mechanism provides approximately twice the holding force compared to a non-jamming rigid pad.

In submerged conditions, although holding forces decreased for all pad variants due to the reduced frictional margins at the water-to-surface interface, the flat jamming pad outperformed the rigid baseline. Specifically, the flat pad achieved 7.41 N under water, while the rigid pad reached only 3.70 N. This indicates that the geometric conformation enabled by the passive particle jamming mechanism provides critical shear resistance and maintains holding even when surface friction is significantly compromised by submersion.

*3) Surface pattern:* The influence of surface texturing was evaluated by comparing the grooved and flat jamming pads. Overall, the grooved pad yielded slightly higher average holding forces than the flat pad in both dry and submerged conditions. In dry settings, the performance difference varied depending on the object geometry. For instance, on cylindrical surfaces like Marker 1, the force improved by 61.8% from 14.46 N (flat) to 23.40 N (groove). However, for highly sloped objects like Marker 12, the performance decreased by 42.4% from 14.51 N (flat) to 8.36 N (groove), as maximizing the conformal contact area becomes more critical than interlocking. Under water, the magnitude of the difference between the two pads diminished; nevertheless, the grooved pad still maintained a marginal average advantage. Therefore, the grooved pattern should be interpreted as a surface-design trade-off that provides average benefits for some object geometries, particularly cylindrical ones, while the flat pad can be preferable for highly sloped objects requiring more continuous conformal contact.

*4) Object-wise performance:* Fig. 6(b) provides a detailed performance breakdown, illustrating how object diameter and slope angle dictate the effectiveness of each design component. Note that Markers 1-4 denote cylindrical objects, while Markers 7-10 and 11-12 represent conical geometries with taper angles of $10^\circ$ and $20^\circ$, respectively.

The passive rotational joint yielded the most significant gains on highly tapered objects, where the transition from unstable point contact to stable surface contact was the primary factor in enhancing holding capability. In contrast, for cylindrical objects, the grooved pattern dominated the performance by enabling geometric interlocking with the surface of the object.

The higher standard deviations observed in specific cases suggest that the failure modes are sensitive to the initial contact configuration and the stochastic nature of local surface interactions between the TPU membrane and the object geometry, especially sudden torque-induced rolling.

*5) Discussion:* The experimental results demonstrate that robust grasping is achieved through the synergy between the passive rotational joint and the particle jamming pad. The passive rotational joint enables the gripper to adapt to conical geometries, ensuring that the pads are oriented to apply normal loads while mitigating detrimental shearing torques on sloped contacts. Once aligned, the particle jamming pad acts as a fine adaptation layer where its variable-stiffness mechanism effectively locks the geometry of the object within the deformed structure of the pad.

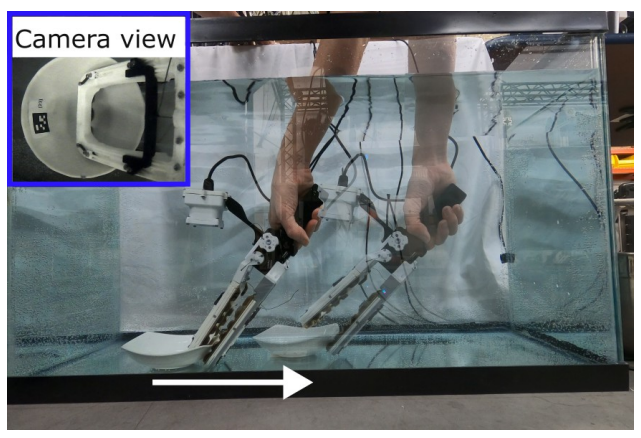

(a) Grasping underwater objects

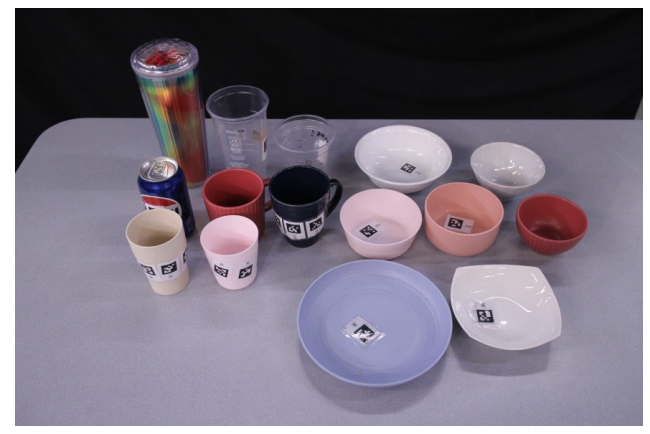

(b) Objects for water experiments

Fig. 7. Underwater experiments setup. A camera inside the waterproof housing can detect markers on the surface of objects to track the movement of the objects.

TABLE II
OBJECT POSE AND SPEED CHANGE UNDER WATER

| Type | Item | Position (mm) | Speed (mm/s) | Orientation (deg) | Angular speed(deg/s) |
|---|---|---|---|---|---|
| Conical object | Blue cup | 5.0 | 0.8 | 9.7 | 1.6 |
| | Red cup | 4.6 | 0.9 | 10.5 | 2.1 |
| | Pink cup | 7.5 | 0.9 | 17.6 | 2.1 |
| | Yellow cup | 8.8 | 1.2 | 24.7 | 3.2 |
| | Large plastic cup | 3.2 | 0.4 | 6.8 | 0.8 |
| | Small plastic cup | 11.0 | 1.5 | 29.0 | 4.2 |
| | Tumbler | 1.9 | 0.4 | 11.2 | 1.8 |
| | Can | 3.5 | 0.5 | 12.9 | 2.0 |
| Plate & Bowl | Large dish | 13.3 | 2.7 | 6.6 | 1.3 |
| | Small dish | 11.9 | 3.2 | 18.5 | 5.1 |
| | Large white bowl | 9.1 | 2.0 | 16.2 | 3.3 |
| | Small white bowl | 3.7 | 1.0 | 3.6 | 0.9 |
| | Orange bowl | 6.5 | 1.2 | 48.1 | 10.6 |
| | Red bowl | 2.5 | 0.5 | 30.6 | 5.9 |
| | Frying pan | 3.2 | 0.7 | 14.7 | 3.8 |

The bimodal performance trend observed in the surface pattern tests highlights a trade-off between geometric interlocking and conformal contact area. While grooves can provide a mechanical advantage on cylindrical surfaces through interlocking, they may reduce the effective continuous contact area on highly sloped conical shapes, where uniform pressure distribution is more critical. Thus, the effect of the groove pattern should be interpreted as object-dependent, with its benefit depending on object geometry and contact conditions.

Ultimately, combining self-alignment with variable-stiffness jamming compensates for the reduced frictional margins inherent in underwater manipulation. This integration provides a versatile and robust solution for the stable grasping of diverse geometries in challenging submerged conditions.

## C. Underwater Item Grasping

The grasping performance of the proposed gripper was evaluated with real-world kitchen items in a submerged environment. Fig. 7(a) shows the pick-and-place protocol: each object was placed in the water tank, grasped, moved to the tank center, and released. Grasp stability was monitored by tracking an AprilTag attached to the object.

For safety and repeatability, the gripper was manually operated using a handle while the self-alignment mechanism and the internal particle jamming pad functioned passively. As shown in Fig. 7(b), the test set included sloped kitchenware as well as plates and bowls to assess versatility. For buoyant items, such as plastic cups and light dishes, the gripper also completed transport while the objects were floating, indicating robust holding under hydrodynamic disturbances.

Object pose and velocity changes were calculated over the manipulation, from the time the gripper closed and reached a steady grasp to the time it opened for release. When the marker was temporarily lost due to reflections, the pose was linearly interpolated between the last valid detection and the next re-detection. Each object was tested over five trials, and we report the mean pose and velocity changes, where both increases and decreases in pose were accumulated in magnitude since reduced motion can also reflect slip recovery or re-alignment.

Table II reports the measured pose change and slip rates during submerged manipulation. Overall, the gripper maintained stable grasps across the tested kitchenware, while the pose change exhibited clear dependency on object geometry and buoyancy. Conical objects provided large contact areas along the outer surface, enabling the jamming pad to conform and suppress translational slip; correspondingly, rigid cylindrical or conical items such as the metal can and tumbler remained within 5 mm positional displacement with minimal average slip speeds of under 1.5 mm/s.

In contrast, plates and bowls were grasped primarily on thin walls with mixed convex and concave curvatures, which limited effective contact area and led to relatively larger positional shifts and faster slip rates, for example, 11-14 mm at 2.0-3.2 mm/s for the large and small dishes. Buoyancy further influenced rotational stability: dense items, such as the frying pan and small white bowl, maintained highly stable orientations with changes under 15 deg and rotational speeds below 4 deg/s, whereas lightweight plastic objects, such as the plastic bowls and small cups exhibited larger rotational variations, up to 48 deg, and peaked at rotational speeds exceeding 10 deg/s as they naturally tended to pivot under buoyant forces and hydrodynamic drag during transport.

Despite these pose changes and slip dynamics, all objects were moved successfully without a single dropping failure across all trials. This outcome suggests that the passive rotational joint provides macroscopic self-alignment while the particle jamming pad enhances shear resistance, together enabling robust underwater grasping even under constrained contact conditions, slippery surfaces, and multi-directional fluidic disturbances.

## IV. CONCLUSION

In this paper, we presented a waterproof passive adaptive gripper designed to tackle the profound challenges of grasping everyday objects in low-friction, underwater environments. To address the frequent grasp failures caused by torque-driven rolling and tangential shear slip on sloped or asymmetric items, our design combines local and global adaptability. This was achieved by combining global self-alignment through a passive rotational joint with local variable-stiffness change via particle-filled pads. The joint mitigates rolling on sloped contacts, while the pad improves shear robustness under compression. This integrated performance is achieved without external vacuum actuation.

Extensive evaluations in both dry and submerged conditions validated that this approach achieves superior performance compared to conventional rigid and fixed-joint baselines. By reducing external torques through the passive rotational joint, maximizing contact friction via the specialized surface patterns, and increasing shear-resistant holding capability through the passive particle jamming mechanism, the proposed gripper demonstrated a remarkable improvement in robustness and stability across diverse kitchenware configurations. Furthermore, the comprehensive waterproof design not only ensured reliable submerged actuation but also enabled an integrated camera to successfully capture a clear view for in-hand object monitoring under water.

Future work will focus on integrating closed-loop grasp regulation using in-hand visual or tactile feedback to detect and mitigate incipient slip during autonomous underwater manipulation. This will improve grasp stability across a wider range of object sizes and slope angles, ultimately enabling contact-rich underwater tasks such as retrieving items from cluttered sinks or dishwashers.

## ACKNOWLEDGMENT

This work was partially supported by Toyota Research Institute (TRI).